\documentclass{article}

\usepackage{microtype}
\usepackage{graphicx}
\usepackage{subfigure}
\usepackage{subcaption}
\usepackage{booktabs} 

\usepackage{hyperref}
\usepackage{amsmath}
\usepackage{amssymb}
\usepackage{amsthm}
\usepackage{algorithm}
\usepackage{algorithmic}
\usepackage{duckuments}

\usepackage{color}

\newcommand{\prac}[2]{\frac{\partial #1}{\partial #2}}
\newcommand{\pw}{\pi_\theta(y_w\mid x)}
\newcommand{\pl}{\pi_\theta(y_l\mid x)}
\newcommand{\probi}{\pi_\theta(y_i\mid x)}
\newcommand{\probj}{\pi_\theta(y_j\mid x)}

\newtheorem{theorem}{Theorem}
\newtheorem{lemma}{Lemma}
\newtheorem{definition}{Definition}
\newtheorem{corollary}{Corollary}
\newtheorem{proposition}{Proposition}
\newtheorem{assumption}{Assumption}

\usepackage[preprint]{icml2024}

\icmltitlerunning{Gradient Imbalance in Direct Preference Optimization}

\begin{document}

\twocolumn[
\icmltitle{Gradient Imbalance in Direct Preference Optimization}

\icmlsetsymbol{equal}{*}

\begin{icmlauthorlist}
\icmlauthor{Qinwei Ma}{yyy}
\icmlauthor{Jingzhe Shi}{yyy}
\icmlauthor{Can Jin}{xxx}
\icmlauthor{Jenq-Neng Hwang}{www}
\icmlauthor{Serge Belongie}{zzz}
\icmlauthor{Lei Li}{www,zzz}

\end{icmlauthorlist}

\icmlaffiliation{xxx}{Rutgers University}
\icmlaffiliation{yyy}{IIIS, Tsinghua University}
\icmlaffiliation{www}{University of Washington}
\icmlaffiliation{zzz}{University of Copenhagen}

\icmlcorrespondingauthor{Qinwei Ma, Lei Li}{qinweimartin@gmail.com,lilei@di.ku.dk}

\icmlkeywords{RLHF, DPO, Machine Learning, Theory}

\vskip 0.3in
]

\printAffiliationsAndNotice{}

\begin{abstract}


Direct Preference Optimization (DPO) has been proposed as a promising alternative to Proximal Policy Optimization (PPO) based Reinforcement Learning with Human Feedback (RLHF). 
However, empirical evaluations consistently reveal suboptimal performance in DPO compared to common RLHF pipelines.
In this work, we conduct a systematic analysis of DPO's training dynamics and identify gradient imbalance as a critical limitation. We demonstrate theoretically and empirically that this imbalance perturbs optimization trajectories, destabilizes learning, and induces suboptimal convergence. 
To address this issue, we propose Balanced-DPO, a simple yet effective modification to the DPO objective that introduces a computationally efficient gradient reweighting mechanism. 
Our experiments demonstrate the effectiveness of Balanced-DPO, validating the theoretical findings and confirming that addressing gradient imbalance is key to improving DPO's performance, highlighting a promising direction for future research.


\end{abstract}

\section{Introduction}

Large Language Models (LLMs) have demonstrated impressive capabilities through pre-training on vast textual datasets \cite{llmfewshotlearner, openai2024gpt4technicalreport, touvron2023llama2openfoundation, anil2023palm2technicalreport}. To align these models with human values and preferences for practical applications, fine-tuning with human feedback is essential \cite{agrawal2024languagemodelsknowtheyre, kadavath2022languagemodelsmostlyknow}. Reinforcement Learning from Human Feedback (RLHF) is a prominent approach for this alignment \cite{ouyang2022traininglmintruction, stiennon2020summarizewithhumanfeedback, ziegler2020finetuninglanguagemodelshuman}, with Proximal Policy Optimization (PPO) being the primary method due to its stability and efficiency \cite{zheng2023secretsrlhflargelanguage}. PPO employs an actor-critic framework \cite{konda1999actor}, constructing a reward model based on human preferences and fine-tuning the policy accordingly. Despite its effectiveness, PPO’s reward-based approach entails a complex training pipeline, necessitating both reward modeling and reinforcement learning components. To address these complexities, Direct Preference Optimization (DPO) \cite{rafailov2024direct} was introduced as a reward-free alternative. Unlike PPO, DPO directly optimizes the model using human preference feedback without requiring a separate reward model, thereby enhancing accessibility and efficiency for certain applications.

However, while DPO shows promise in specific tasks, its performance diminishes in more complex scenarios such as code generation or nuanced decision-making, where subtle human preferences are critical. Previous studies \cite{xu2024dposuperiorppollm} have identified several challenges limiting DPO’s broader applicability:
\begin{enumerate}
    \item \textbf{Sensitivity to Data Quality:} DPO relies heavily on high-quality preference data, which can be noisy or inconsistent in real-world settings.
    \item \textbf{Vulnerability to Distribution Shifts:} DPO struggles when there is a discrepancy between training and deployment data distributions, leading to degraded performance in novel or unseen scenarios.
    \item \textbf{Overestimation of Out-of-Distribution (OOD) Responses:} DPO tends to overestimate the probabilities of OOD responses, particularly under distribution shifts, posing risks for safe and robust deployment.
\end{enumerate}

While existing work has made progress in addressing the first two challenges, the third challenge—overestimation of OOD responses—remains largely underexplored. This limitation can lead to unsafe or unpredictable model behavior, especially in complex or safety-critical tasks. Moreover, most previous works have failed to provide a clear analysis of why DPO faces these challenges, particularly given that it shares the same intrinsic reward as PPO.

In this paper, we identify \textit{gradient imbalance} during training as the root cause of DPO’s limitations. Unlike PPO, which maintains a balanced gradient update between winning and losing responses, DPO exhibits an imbalance that disproportionately updates model weights based on losing responses. Specifically, for each pair of winning and losing responses, the probability update is more focused on the losing responses compared to PPO, which intuitively leads to the overestimation of OOD responses. The gradient imbalance also amplifies sensitivity to data quality and vulnerability to distribution shifts. These challenges collectively hinder DPO’s effectiveness, particularly in complex tasks where subtle human preferences are critical.

To address the gradient imbalance in DPO, we propose a novel method, \textbf{Balanced-DPO}. By ensuring a more balanced update between winning and losing responses, Balanced-DPO achieves: 1) enhanced alignment with human preferences, 2) improved robustness to distribution shifts, and 3) reduced overestimation of OOD responses. We verify its effectiveness on various benchmarks, thereby demonstrating that it is a promising approach to mitigating gradient imbalance and improving DPO and other algorithms based on pairwise feedback.

We summarize our core contributions as follows: 
\begin{enumerate} 
    \item \textbf{Theoretical Analysis of Gradient Imbalance:} We provide a novel and comprehensive theoretical analysis of DPO, identifying gradient imbalance as the root cause of its performance limitations, including sensitivity to data quality, vulnerability to distribution shifts, and overestimation of out-of-distribution (OOD) responses. 
    \item \textbf{Insights into Learning Dynamics:} Unlike prior work that primarily focuses on empirical observations or the static properties of DPO, we analyze its learning dynamics, shedding light on how the imbalance in gradient updates between winning and losing responses adversely impacts training. 
    \item \textbf{Validation of Theoretical Insights through Balanced-DPO:} Based on our theoretical findings, we propose Balanced-DPO as a concrete method to address gradient imbalance. This method is specifically designed to validate our theoretical insights and demonstrate the practical benefits of correcting gradient imbalance in pairwise-feedback-based optimization methods. 
\end{enumerate}

\section{Related Works} \label{section:related_work}

In this section, we review key works that: 1) propose significant variants of DPO, or 2) analyze PPO, DPO, or their variants from theoretical or empirical perspectives.

First, considering variants of DPO, the most important ones are IPO \cite{azar2023ipo}, CPO \cite{xu2024cpo}, and KTO \cite{ethayarajh2024kto}. These works first analyze DPO from different perspectives and then propose new loss functions to improve its performance. Among them, KTO introduces an option to apply different weights to winning and losing responses, which aligns with our analysis. However, it applies equal weights during evaluation and does not provide further analysis on this aspect. Besides these, other notable methods include $\alpha$-DPO\cite{wu2024alpha}, $\beta$-DPO\cite{wu2024beta}, rDPO \cite{chowdhury2024rdpo}, iterative DPO\cite{xiong2024iterative}, etc.

For works focusing on analysis, the most notable is \cite{xu2024dposuperiorppollm}, which compares PPO and DPO, analyzing and empirically verifying the drawbacks of DPO. \cite{saeidi2024insights} provides a thorough comparison of various DPO variants alongside SFT. \cite{lee2024toxic} constructs a toxicity dataset to study how DPO reduces output toxicity, offering insights into its learning process.

\section{Preliminaries} \label{section:theory}

\subsection{Formulation} \label{section:prelim}

In Section~\ref{section:theory}, all analyses are conducted under a fixed input \( x \), meaning that all probabilities, model outputs, and data samples are assumed to share the same input or context \( x \).   

Let the total response space be \( \mathcal{A} \), with a subset covered by the training dataset denoted as \( \mathcal{A}_T \). The training dataset is \( \mathcal{D}_T \), where each sample can take one of two forms. The classic preference pair format is \( (y_1 \succ y_2) \), indicating that \( y_1 \) is preferred over \( y_2 \). However, for a more simplified and unified derivation in this section, we also adopt an alternative form, \( (y_1, y_2, \tau) \). Here, \( \tau = \pm 1 \) represents the preference order between the two responses: \( \tau = 1 \) indicates \( y_1 \succ y_2 \), while \( \tau = -1 \) indicates the opposite.

With the above simplifications, the original DPO loss presented in \cite{rafailov2024direct} could be written as:

\begin{multline}
    \label{dpoloss}
    \mathcal{L}_{dpo} = -\mathbb{E}_{x\sim \mathcal{D}(x), y\sim \mathcal{D}(y\mid x)}[\log\sigma(\beta\log(\frac{\pi_\theta(y_w\mid x)}{\pi_{ref}(y_w\mid x)}) \\ -\beta\log(\frac{\pi_\theta(y_l\mid x)}{\pi_{ref}(y_l\mid x)}))]
\end{multline}


Although symmetric in form for $y_w$ and $y_l$, the gradient norms differ. Specifically:

\begin{equation} 
\label{equation:dpograd}
\frac{\partial \mathcal{L}_{dpo}}{\partial \pi_\theta(y_w\mid x)} \Big/ \frac{\partial \mathcal{L}_{dpo}}{\partial \pi_\theta(y_l\mid x)} = -\frac{\pi_{\theta}(y_l\mid x)}{\pi_\theta(y_w\mid x)} 
\end{equation}

We may compare it with PPO. The PPO process includes two stages, reward model training and policy model training. The loss for reward model training is:

\begin{equation}
    \mathcal{L}_R(r_\phi) = \mathbb{E}_{(x,y_w,y_l)\in \mathcal{D}}[\log\sigma(r_\phi(x,y_w)-r_\phi(x,y_l)]
\end{equation}

It is easy to see that $|\prac{\mathcal{L}_{R}}{r_\phi(x,y_w)}| = |\prac{\mathcal{L}_{R}}{r_\phi(x,y_l)}|$. For the loss for PPO policy, according to \cite{ouyang2022traininglmintruction} is:

\begin{equation}
    \mathcal{L}_{ppo}(\pi_\theta) = \mathbb{E}_{x \sim p_{\text {data }}, y \sim \pi_\theta}[r(x, y)-\beta \log \frac{\pi_\theta(y \mid x)}{\pi_{\mathrm{ref}}(y \mid x)}]
\end{equation}

The gradient of PPO policy can be expressed as follows.

\begin{equation}
    \prac{\mathcal{L}_{ppo}}{\pi_\theta(y\mid x)} = \mathbb{E}_{x \sim p_{\text {data }}}[r_\phi(x, y)-\beta \log \frac{\pi_\theta(y \mid x)}{\pi_{\mathrm{ref}}(y \mid x)} - \beta]
\end{equation}

Since an optimal reward model for PPO $r_\phi^*$ should satisfy: \cite{Sanghi2024} $r_\phi(x, y)\propto \log(P_\mathcal{D}(y\mid x))$, we can see that PPO tends to focus more on the winning responses, which is very different from DPO.










This comparison raises a fundamental question: does the imbalanced gradient in DPO aid or hinder learning? Could it contribute to DPO's known drawbacks?

\subsection{Assumptions}
\label{section:assumptions}

To formalize the problem and focus on the learning process, we wish to minimize the effect caused by the difference in model categories. We assume the outputs of the model are the raw logits, denoted as $s_i$ for each response $i\in R$, and the probability of a response $i$ should be 

\begin{equation}
    \label{eq:logitandprob}
    p_i = \frac{e^{s_i}}{\sum\limits_{j\in \mathcal{A}} e^{s_j}}
\end{equation}

Our analysis is mainly based on the following assumptions:

\begin{assumption}
    \label{assumption1}
    We assume that $\|\prac{s_i}{\theta}\|$ is similar for all $i$, and approximate its value with $g$. 
\end{assumption}

\begin{assumption}
    \label{assumption2}
    Let $c_{ij} = \frac{\prac{s_i}{\theta} \cdot \prac{s_j}{\theta}}{\|\prac{s_i}{\theta}\| \cdot \| \prac{s_j}{\theta}\|}$, then we assume that
    \begin{equation}
        c_{ij} \approx 0, ~\forall i\neq j
    \end{equation}
\end{assumption}

Assumption~\ref{assumption1} is made only for simplicity, as it is generally impossible to analyze the difference in gradient values for different logits, and out of symmetry, it is reasonable to suggest that they are similar.

Assumption~\ref{assumption2} indicates that the gradient updates for different logits are almost orthogonal with each other. The major reason to adopt this assumption is for simplicity. There are several evidences justifying the claim.

1) If $\prac{s_i}{\theta}$ and $\prac{s_j}{\theta}$ are two random vectors the space, then $\mathbb{E}_{i,j}[c_{ij}] = 0$ and $\mathbb{E}_{i,j}[c_{ij}^2]=\mathcal{O}(\frac 1 N)$, where $N$ is the number of model parameters. That is, for large enough models, the correlation would be small enough to be neglected.

2) Empirically, we found that $\mathbb{E}_{i,j}[c_{ij}] < 0$. We provide a brief analysis in Appendix~\ref{appendix:negativecorr} that, such a negative correlation will further enhance our result.

3) Generally speaking, if $\|\mathbb{E}_{i,j}[c_{ij}]\|$, this implies that there's much redundancy in the model, which either indicates that the model is larger than needed, or that the model is not well optimized. 

\section{Theory}

In this section, we delve deeper into the dynamics in each training step to see what actually happens in an epoch, Here we consider the update given a certain pairwise sample $(i, j, \tau_{ij})$, and analyze how different choices of the loss function affect the overall performance.

\subsection{General Analysis}

In this section, w.l.o.g. we let $\beta=1$ in the DPO loss for simplicity.

\begin{definition}
    We say an preference loss $\mathcal{L}$ is \textbf{balanced} if:
    \begin{equation}
        \mathbb{E}[\prac{\mathcal{L}}{\pw} + \prac{\mathcal{L}}{\pl}] = 0
    \end{equation}
    else we say the loss is \textbf{imbalanced}. In this case, if $|\prac{\mathcal{L}}{\pw}| > |\prac{\mathcal{L}}{\pl}|$, we say it is \textbf{positively imbalanced}, else it is \textbf{negatively imbalanced}.
\end{definition}

By definition, we can easily find that:

\begin{corollary}
    Standard DPO loss is \textbf{negatively imbalanced}.
\end{corollary}

\begin{corollary}
    PPO reward loss is \textbf{balanced}.
\end{corollary}

As for PPO policy loss, if we consider two samples as a 'pair' we may extend the definition to suit it. In this case we can claim that:

\begin{corollary}
    PPO policy loss is (by broad definition) \textbf{positively imbalanced} with a good reward model.
\end{corollary}

Next, we may analyze how the imbalance affect the learning process. We assume the loss function satisfy the form $\mathcal{L} = -\log\sigma(f(\pw,\pl))$, and we let $\kappa_{ij}$ denote $e^{-\mathcal{L}}$ for a pairwise sample $(i, j,\tau_{ij})$, i.e. the $\sigma(\cdot)$ term in the loss function, and let $\eta$ denote the learning rate. 

\begin{lemma}
    \label{lemma:pracps}
    \begin{equation}
        \prac{p_i}{s_j} = p_i(\delta_{ij} - p_j)
    \end{equation}
    where $\delta(\cdot)$ is the Kronecker delta function.
\end{lemma}

\begin{theorem}
    \label{theorem:deltajsi}
    If we use DPO to update a model on a sample $(i,j,\tau_{ij})$, assume $\beta=1$, then the change in $s_i$ with this sample would be:
    \begin{equation}
        \Delta^js_i = \eta g^2 \tau_{ij}(1-\kappa_{ij}) 
    \end{equation}
    (Recall that $g=\|\prac{s_i}{\theta}\|$)
    For \textbf{balanced} losses,
    \begin{equation}
        \Delta^js_i = C\eta g^2 \tau_{ij}(1-\kappa_{ij})(1- \probi+\probj)
    \end{equation}
    where $C$ is a constant. (We omit $C$ in the following sections, as it does not have an effect after normalization.
\end{theorem}

Next, we delve into the probability update of each response.

Let $w_i = \sum\limits_{(i, j, \tau_{ij})\in \mathcal{A}}\tau_{ij}(1-\kappa_{ij})$, and $\gamma=\eta g^2$, then the probability update after a training epoch for both cases would be as follows.

\begin{theorem}
    \label{theorem:probupdate}
    We assume all responses in the dataset are drawn from a uniform distribution. Then, for DPO,
    \begin{equation}
        \Delta \probi = \gamma \probi (w_i - \sum\limits_{j\in A}w_j\probj) 
    \end{equation}
    For an algorithm with a \textbf{balanced} loss, if we assume the response space is large and the model has not learnt well enough, so that $\probi$ is relatively small for any $y_i\in \mathcal{A}$, so that such an approximation holds:
    \begin{equation}
        1-\probi+\probj \approx 1, \forall y_i,y_j 
    \end{equation}
    Then we have:
    \begin{multline}
        \Delta \probi \approx C\gamma \probi (w_i\probi \\ - \sum\limits_{j\in A}w_j\probj^2) 
    \end{multline}
    where $C$ is a constant.

    In the following parts, we assume $C=1$ as it makes no effect after normalization.
\end{theorem}

If we look at the intuition of this formula, $\gamma \probi$ is the same coefficient for any algorithm, and the last terms in both formulas serve as a weighted averaging to make $\sum\limits_{y_i\in \mathcal{A}}\Delta \probi = 0$. Hence, we can intuitively claim that the distribution of $w_i$ and $w_i\probi$ results in some key performance differences between DPO and some other algorithms. Here we mostly compare DPO with algorithms with a \textbf{balanced} loss, to which category the PPO reward loss falls in. However, the following analysis can be easily extended to \textbf{positively imbalanced} losses including the loss used for policy learning in PPO. 

\subsection{Synthetic Simulation}

\begin{figure}[ht]
    \centering
    \includegraphics[width=\columnwidth]{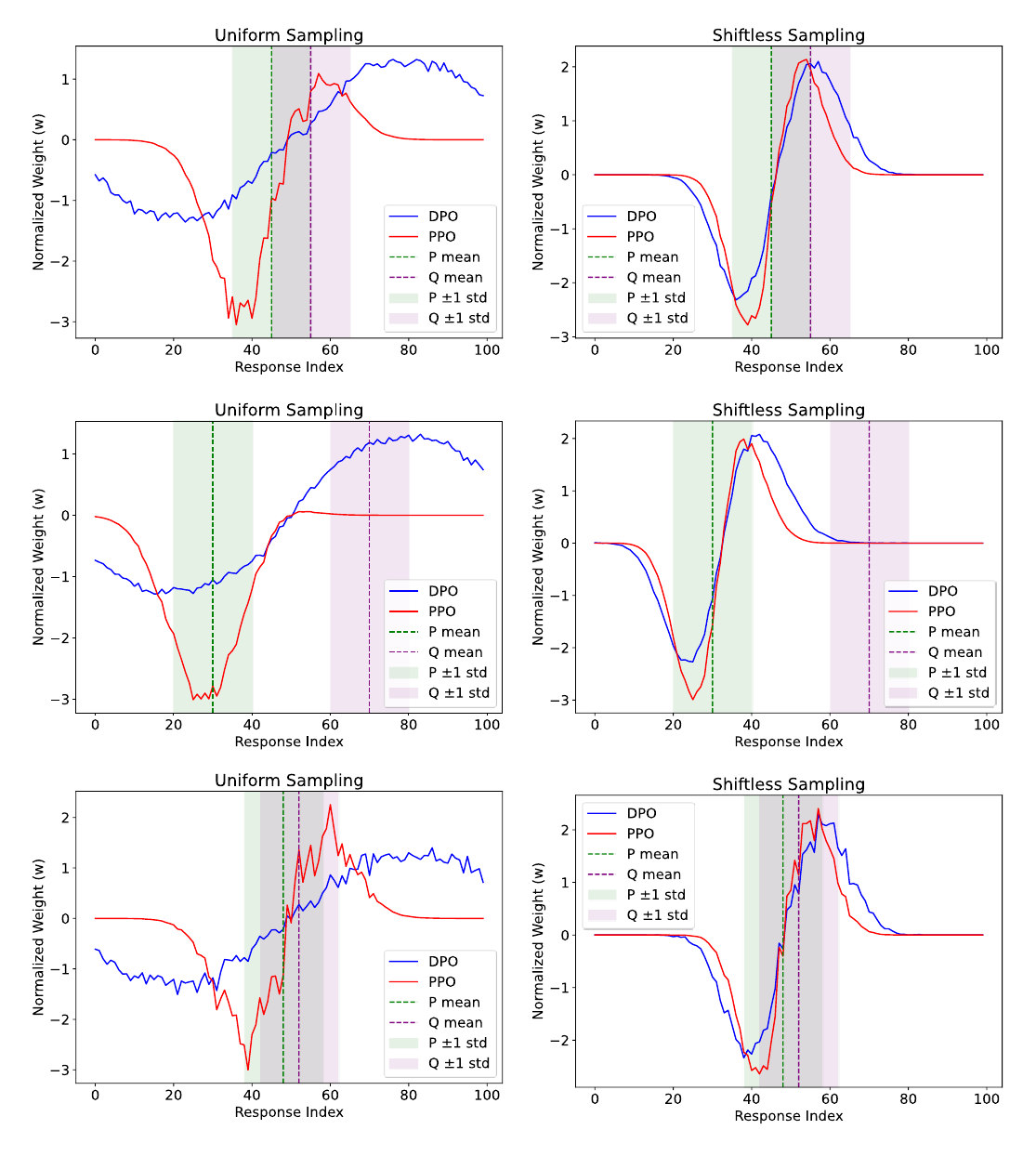}
    \caption{The distribution of $w_i$ (for DPO) and $w_i\pi_\theta(y_i\mid x)$ (for PPO) in different scenarios. In each figure, the red curve is for PPO, and the blue curve is for DPO; the two colored regions show the $\mu$ and $\sigma$ for both P and Q distributions, where the green region stands for P and the violet region stands for Q. The images in the left column use uniform sampling, while the images in the right column sample according to the current model distribution, which is to simulate the case without any distribution shift. In all figures, the response set is $\mathcal{A}=\{0,1,\cdots,99\}$. In the first row, $\mu_P=45,~\mu_Q=55$; in the second row, $\mu_P=30,~\mu_Q=70$; in the third row, $\mu_P=48, \mu_Q=52$; in all figures, $\sigma_P^2=\sigma_Q^2=100$.}
    \label{fig:gaussianresults}
\end{figure}

Recall that in \cite{xu2024dposuperiorppollm}, three challenges were listed for DPO. We first do a simple simulation for the distribution of $w_i$ (for DPO) and $w_i\pi_\theta(y_i\mid x)$ (for PPO-like methods), assuming that the model output distribution (denoted as the \textbf{P distribution}) is a Gaussian distribution, i.e.
\begin{equation}
    \pi_\theta(y\mid x) = \frac{1}{\sqrt{2\pi\sigma_p}}e^{-\frac{(y-\mu_p)^2}{2\sigma_p^2}}
\end{equation}
and we assume that the preference is given by a Gaussian-like utility function (denoted as the \textbf{Q} distribution), more specifically:
\begin{equation}
    u(x,y) = \frac{1}{\sqrt{2\pi\sigma_u}}e^{-\frac{(y-\mu_u)^2}{2\sigma_u^2}}
\end{equation}
and in a dataset, the preference is labeled according to
\begin{equation}
    P(y_1\succ y_2\mid x) = \frac{u(x,y_1)}{u(x,y_1)+u(x,y_2)}
\end{equation}
In our simulation, we always set $\sigma_p=\sigma_u=\sigma$ for simplicity. We simulate with two sampling methods, the corresponding distribution denoted as the \textbf{D distribution}. For \textbf{uniform} sampling method $P_U(y_1,y_2\mid x) = \text{const}$, which indicates that all responses have equal probabilities to be sampled; on the other hand, for the \textbf{shiftless} method $P_s(y_1,y_2\mid x)\sim \pi_\theta(y_1\mid x)\pi_\theta(y_2\mid x)$, which is the case where there is no distribution shift between the model output distribution (P distribution) and dataset distribution (D distribution). A visualized comparison is shown in Figure~\ref{fig:gaussianresults}, where the representatives of three different scenarios are displayed, $|\mu_p-\mu_u| \approx \sigma$ (first row), $|\mu_p-\mu_u|\gg \sigma$ (second row) and $|\mu_p-\mu_u|\ll \sigma$ (third row).

We can see from the curves that, when using uniform sampling, which resembles the case that the distribution shift between the model and the dataset is large, the distribution of $w_i$ for DPO will also suffer from a large distribution shift from both the model distribution and the dataset distribution, while PPO does not encounter such a problem. However, if there is no distribution shift, i.e. the dataset distribution is the same as the model distribution (which is common in online learning), DPO and PPO's performances are similar. Also, we can see the distribution for PPO is more 'concentrated', so that the updates for OOD responses are significantly smaller compared to DPO.

In the following subsections, we analyze how the distribution difference of $w_i$ and $w_i\pi_\theta(y_i\mid x)$ is related to the challenges DPO is facing respectively. 

For simplicity, in the following sections we adopt the following symbols: 
\begin{align*}
    p_i &= \probi \\
    \alpha_i &=\sum\limits_{j\neq i} \frac{p_i}{p_i+p_j}\\
    \beta_i &= \sum\limits_{j\neq i} \frac{e^{r^*_i}}{e^{r^*_i}+e^{r^*_j}}
\end{align*}

\subsection{Detailed Analysis}

\subsubsection{Distribution Shift}

The first challenge is that, DPO is more sensitive to distribution shifts than PPO. We claim that this actually originates from the distribution shift in $w_i$. We can formalize this into the following.

\begin{theorem}
    \label{theorem:distshift}
    Let $y_i$ lie in a one-dimensional space. Suppose $\pi_\theta(y\mid x)$ has global maximum at $y=\mu_p$ and decreases \textbf{super-exponentially}. Suppose the optimal reward model $r^*$ has global maximum at $y=\mu_q$ and decays \textbf{super-linearly}. W.l.o.g we assume $\mu_p\leq \mu_q$. If there is a distribution shift, i.e. $\mu_p\neq \mu_q$, then:
    \begin{itemize}
        \item If $y^*$ is a local extremum for $w(y_i)$, then $y^*>\mu_q$ or $y^*<\mu_p$.
        \item If $y^*$ is a local extremum for $w(y_i)$ and $y^*>\mu_q$, then exists a local extremum $\hat{y}^{*}$ for $w(y_i)\pi(y_i \mid x)$ satisfying $\mu_p<\hat{y}^{*}<y^*$.
        \item If $y^*$ is a local extremum for $w(y_i)$ and $y^*>\mu_q$, then exists a local extremum $\hat{y}^{*}$ for $w(y_i)\pi(y_i \mid x)$ satisfying $y^*<\hat{y}^{*}<\mu_p$.
    \end{itemize}
\end{theorem}

An intuitive conclusion from Theorem~\ref{theorem:probupdate} and Theorem~\ref{theorem:distshift} is that, if the dataset is sampled uniformly from a response space, which is an extreme case for distribution shift, then update of DPO would be more biased. The visualized result in Figure~\ref{fig:gaussianresults} shows that, sometimes the extremum of $w_i$ might deviate very far from the centers of the two distributions, which makes the problem stated in Theorem~\ref{theorem:distshift} even more severe.

\subsubsection{Data Quality}
It is widely acknowledged that DPO is more sensitive to data quality than PPO. The primary reason is that PPO first learns a reward model, which ensures a certain level of generalization. However, this may not be the only reason. Taking a different perspective on this problem, we show that gradient imbalance could also be a contributing factor in two distinct aspects.

First, we demonstrate that \( w_i \) follows a less \emph{concentrated} distribution than \( w_i p_i \), leading to a noisier update. More specifically, we show that it exhibits a larger variance. Generally speaking, if we control the 'scale' of two distributions to be the same, the one with greater variance is more dispersed. However, since \( p_i \) is small, we must \textbf{normalize} \( p_i \) by enforcing \( \mathbb{E}[p_i^2] = 1 \) to eliminate the impact of its low absolute value.

\begin{lemma}
    \label{lemma:wi_simplify}
    \begin{equation}
        w_i = - (\alpha_i-\beta_i), ~\forall i
    \end{equation}
\end{lemma}

\begin{lemma}
    \label{lemma:fym}
    Assume $X$ is a random variable and $f,g:\mathbb{R}\to\mathbb{R}^+$ are bounded increasing functions, then
    \begin{equation}
        \operatorname{Var}[f(X)g(X)]\geq \operatorname{Var}[f(X)]\mathbb{E}[g(X)^2]
    \end{equation}
\end{lemma}

\begin{theorem}
    \label{theorem:variance}
    If $\beta_i$ monotonically increases with $p_i$, then
    \begin{multline}
        \operatorname{Var}[w_ip_i] - \operatorname{Var}[w_i]\mathbb{E}[p_i^2] \geq 2(\operatorname{Cov}[\alpha_i,\beta_i]\mathbb{E}[p_i^2] \\ - \operatorname{Cov}[\alpha_ip_i, \beta_ip_i])
    \end{multline}
\end{theorem}

Notice that in the previous section, we discussed how DPO suffers from distribution shift, and as we are trying to discuss the problem in another aspect, it is natural here to control the distribution shift small, hence $\beta_i$ monotonically increases with $p_i$ is a reasonable assumption. Also, this lower bound is pretty loose because Lemma~\ref{lemma:fym} is a bound for very general cases.

Generally, this indicates that PPO updates the probabilities in a more 'centralized' manner. Along with previous analysis that PPO has less update bias compared to DPO, this ensures that PPO updates the probabilities in a more stably compared with DPO. 

On the other hand, we consider the case where there is a small perturbation to each sample in the dataset, and show that DPO will be more affected by such a perturbation compared with PPO.

\begin{theorem} 
    \label{theorem:dataquality}
    Suppose the preference of $y_i\succ y_j$ is given by $r^*$ with a small perturbation, i.e. for every sample $P(y_i\succ y_j) =\frac{r^*_i}{r^*_i+r^*_j} + \epsilon_i$ where $\epsilon$ is a small random variable with $\mathbb{E}[\epsilon_i]=0$. Let $w^*$ denote the $w$ value without perturbation, then for any $\mathcal{A}_0\subseteq\mathcal{A}$ satisfying:
    \begin{equation}
        \mathbb{E}_{y_i\in \mathcal{A}_0} [p_i^2] \geq \mathbb{E}_{y_i\in\mathcal{A}}[p_i^2],
    \end{equation}
    we have (notice that all variances are taken over $y_i\sim \mathcal{A}_0$)
    \begin{equation}
        (\operatorname{Var}[w_i]-\operatorname{Var}[w^*_i])\mathbb{E}_{y_i\sim \mathcal{A}
        }[p_i^2] \leq \operatorname{Var}[w_ip_i] - \operatorname{Var}[w^*_ip_i]
    \end{equation}
\end{theorem}

Recall that the $\mathcal{A}_0$ here can also be the whole response set $\mathcal{A}$. This indicates that, the 'concentration' is better kept in PPO than DPO, implying that DPO is more sensitive to data quality than PPO. 

\subsubsection{OOD Responses}

\label{section:oodresponses}

Another challenge faced by DPO, according to \cite{xu2024dposuperiorppollm}, is that it tends to overestimate the OOD responses compared with PPO, which originates from the offline nature of DPO. As it has seldom been analyzed by previous works, and also it is not commonly discovered in other works, we take a closer look at the synthetic dataset constructed in the original paper to discuss why this happens and how it might be related with the gradient imbalance.

Actually the figure demonstrated in the paper shows that the PPO reward model does not avoid the OOD overestimation problem, and the advantages comes from the policy learning process. However, the original paper claimed that DPO does not guarantee an OOD response $y_i$ with $\pi_{ref}(y_i\mid x)=0$ to have $\pi_{dpo}(y_i\mid x)=0$, which is false according our analysis. In fact this property is guaranteed by the learning process of DPO.

\begin{theorem}
    \label{theorem:oodnoprob}
    If $\pi_{ref}(y\mid x)=0$ and $(y\mid x)$ is out-of-distribution, then $\pi_{dpo}(y\mid x)=0$.
\end{theorem}

Moreover, we can show that in many cases, DPO even overestimates the OOD responses \textbf{less} than balanced losses.

\begin{lemma}
    \label{lemma:fym2}
    Assume $X$ is a random variable and $f,g:\mathbb{R}\to\mathbb{R}^+$ are bounded increasing functions, then
    \begin{equation}
        \text{Cov}[f(X)g(X),g(X)]\geq \mathbb{E}[f(X)]\operatorname{Var}[g(X)]
    \end{equation}
\end{lemma}

\begin{theorem}
    \label{theorem:oodupdate}
    If $\beta_i$ and $\alpha_i$ are independent, then:
    \begin{equation}
        \sum_{j\in\mathcal{A}} w_jp_j^2 < \sum_{j\in\mathcal{A}} w_jp_j \mathbb{E}_{k\in \mathcal{A}}\left[p_k\right]
    \end{equation}
\end{theorem}

This indicates that, if we only compare the loss term, DPO performs even better on the OOD issue compared to PPO; this indicates that, the OOD issue might not originate from the loss, but more from the fact that PPO is regularized not to be too far apart from the reference model.

Moreover, as we will show in Section~\ref{section:toyexp}, we constructed a synthetic dataset with a Gaussian-distributed preference and masked some specific (prompt, response) pairs. We do not observe apparent OOD issues in the result. This might imply that the OOD issue only existed in that extreme synthetic case where only the diagonal is preferred and \emph{all others} are treated equally, and is not severe in real-world cases.

\section{Method} \label{section:method}

From our analysis, we observe that smoothing the learning process requires balancing the gradient updates of the weights. To demonstrate that the improvement stems directly from gradient balancing, we introduce only minor modifications to the loss term in Equation~\ref{dpoloss}. Specifically, we first propose a simple approach, Naive Balanced DPO (Naive-bDPO), which applies a straightforward adjustment to the loss function.

\subsection{Naive Balanced DPO}

As a naive balancing method, we aim to keep the modification as small as possible but increase the gradient for the winning response. This directly leads to such a loss term for Naive-bDPO:

\begin{equation}
\begin{aligned}
    \mathcal{L}_{bdpo} & = -\log\sigma(\beta\lambda\log(\frac{\pi_\theta(y_w\mid x)}{\pi_{ref}(y_w\mid x)})- \\
    &\beta\log(\frac{\pi_\theta(y_l\mid x)}{\pi_{ref}(y_l\mid x)} - \beta(\lambda-1)\log(\frac{\pi_\theta^{\text{detached}}(y_w\mid x)}{\pi_{ref}(y_w\mid x)})))
\end{aligned}
\end{equation}

where $\pi_\theta^{\text{detached}}$ is equal to $\pi_\theta$ but requires no gradient calculation, and lambda is a gradient imbalance ratio. This is a 'brute force' way to reweight the gradients. Notice that this method does not change the value of the loss but only changes its gradient. This is to show that, even with the same set of optimal solutions, how the model is updated throughout the learning process is also important for analyzing the algorithm.

Theoretically, to make the gradient fully balanced, $\lambda$ should be approximately proportional to $\frac{\pi_\theta(y_w\mid x)}{\pi_\theta(y_l\mid x)}$. But naturally, this will often make the gradient too large for a stable training process. In practice, we found it better to set it in a log-scale manner, even if it does not fully address the imbalance.

\begin{equation}
    \lambda = 1 + clip(\log(\frac{\pi_\theta(y_w\mid x)}{\pi_\theta(y_l\mid x)}), \Delta\lambda_{min}, \Delta\lambda_{max})
\end{equation}

$\Delta\lambda_{min}$, $\Delta\lambda_{max}$ are two hyper-parameters to set for `gradient clipping'. Naturally, we wish to emphasize the positive samples only, which means that $\Delta\lambda_{min}$ should be 0 to make sure the gradient is not diminished when $\pi_\theta(y_w\mid x) < \pi_\theta(y_l\mid x)$. However, another approach is to make the gradient more balanced in any case, where $\Delta\lambda_{min} = \frac{1}{1+\Delta\lambda_{max}} - 1$. The former approach is the \textbf{asymmetric} approach, and the latter is the \textbf{symmetric} approach. 

\subsection{Naive Balanced DPO v2}

Although Naive-bDPO keeps the loss value unchanged, detaching the decreased term makes it unable to converge to a local minimum of the loss. To avoid this problem, we provide another alternative which removes the last term to make it natural and simple. The loss for Naive bDPO v2 is:

\begin{equation}
\begin{aligned}
    \mathcal{L}_{bdpo} & = -\log\sigma(\beta\lambda\log(\frac{\pi_\theta(y_w\mid x)}{\pi_{ref}(y_w\mid x)})- \\
    &\beta\log(\frac{\pi_\theta(y_l\mid x)}{\pi_{ref}(y_l\mid x)}))
\end{aligned}
\end{equation}

Similar to Naive bDPO, Natural bDPO also has two version, \textbf{asymmetric} and \textbf{symmetric}, depending on the clipping for $\lambda$.

\subsection{Balanced DPO}

Naive and Natural bDPO are both very simple alternatives to the standard bDPO. We wish to demonstrate by the effectiveness of these methods that, balancing the gradients is an efficient and promising direction to improve the DPO algorithm. However, the performance improvement with such a simple and brute-force modification is limited. Based on the insights, we propose a new method named Balanced DPO (bDPO) to mitigate the gradient imbalance. The loss for bDPO is:

\begin{equation}
\begin{aligned}
    \mathcal{L}_{bdpo} & = -\log\sigma(\beta\lambda_w\log(\frac{\pi_\theta(y_w\mid x)}{\pi_{ref}(y_w\mid x)})- \\
    &\beta(1-\lambda_w)\log(\frac{\pi_\theta(y_l\mid x)}{\pi_{ref}(y_l\mid x)}))
\end{aligned}
\end{equation}

where 

\begin{equation}
    \lambda_w = \frac{\log\pl}{\log\pw + \log\pl}
\end{equation}

Notice that $\log\pw$ and $\log\pl$ are both negative, so if $\pw>\pl$, then $\lambda_w>\frac 1 2$. 

\begin{proposition}
    When $\pi_\theta \equiv \pi_{ref}$, i.e. in the initial training stage, the gradient satisfies:
    \begin{equation} 
        \frac{\partial \mathcal{L}_{bdpo}}{\partial \pi_\theta(y_w\mid x)} \Big/ \frac{\partial \mathcal{L}_{bdpo}}{\partial \pi_\theta(y_l\mid x)} = -\frac{\pi_{\theta}(y_l\mid x)\log\pl}{\pi_\theta(y_w\mid x)\log\pw} 
    \end{equation}
\end{proposition}

Compared with Equation~\ref{equation:dpograd}, we can see that this approach makes the loss more balanced.

\begin{proposition}
    \label{proposition:monotonic}
    Let
    \begin{align}
        g_w &= \mathbb{E}_{(y_w\succ y_l)\sim \mathcal{D}} \left[\prac{\mathcal{L}_{bdpo}}{\pw}\right] \\
        g_l &= \mathbb{E}_{(y_w\succ y_l)\sim \mathcal{D}} \left[ \prac{\mathcal{L}_{bdpo}}{\pl} \right]
    \end{align}
    then $g_w$ monotonically increases during the training process, while $g_l$ monotonically decreases.
\end{proposition}

By Proposition~\ref{proposition:monotonic}, we can see that, throughout the training process, the gradient imbalance will be gradually diminished as we wish.

\section{Experiment} \label{section:experiment}

\subsection{Synthetic Dataset} \label{section:toyexp}

\subsubsection{Setting}

We first conduct our methods on a most simplified setting, which is an advanced and less extreme version of the toy experiment setting in \cite{xu2024dposuperiorppollm}. We let both the prompt space $\mathcal{S}$ and the response space $\mathcal{A}$ to be discrete and have size 20, labeled as 1-20. For simplicity, we set the most preferred response for each prompt $x\in\mathcal{S}$ to be $y=x$ numerically. We assume that the preference distribution in the dataset satisfies the following.

\begin{equation}
    \mathcal{P}(y_1\succ y_2\mid x) = \frac{u(y_1\mid x)}{u(y_1\mid x)+u(y_2\mid x)}
\end{equation}

where $u(y\mid x)$ denotes the preference utility for each response $y$ given a prompt $x$. In this toy experiment, 

\begin{equation} \label{eq:utility}
    u(y\mid x)\propto e^{-\alpha (y-x)^2}
\end{equation}

Moreover, to simulate the OOD responses, we randomly mask some (prompt, response) pairs, i.e. let them never exist in the training dataset, which can be viewed as OOD samples.

\subsubsection{Result}

The result is shown in Table~\ref{table:toyresult}, where \textbf{nbDPO} stands for \emph{Naive Balanced DPO}, and \textbf{bDPO} stands for \emph{Balanced DPO}. The \emph{mask} indicates the proportion of (prompt, response) pairs that are masked during the training process.

\begin{table}[ht]
    \centering
    \caption{Reward achieved by different algorithms with varying mask rates on the synthetic dataset.}
    \label{table:toyresult}
    \resizebox{0.7\columnwidth}{!}{%
    \begin{tabular}{lccc}
        \toprule
        \textbf{Mask Rate} & \textbf{0.0} & \textbf{0.2} & \textbf{0.4} \\
        \midrule
        baseline       & 4.28  & 3.98  & 4.71  \\
        nbDPO (asymm)  & 4.56  & 4.46  & 5.02  \\
        nbDPO (symm)   & 5.40  & 5.30  & 6.15  \\
        nbDPOv2 (asymm) & 3.79  & 4.05  & 4.04  \\
        nbDPOv2 (symm)  & 4.45  & 4.24  & 4.82  \\
        bDPO           & 4.45  & 4.40  & 4.68  \\
        \bottomrule
    \end{tabular}%
    }
\end{table}

First of all, as shown in Table~\ref{table:toyresult}, the results are promising, as most methods successfully beat the baseline, especially the symmetric nbDPO. Moreover, there are other findings: 1) nbDPO v2 performs worse than v1. This suggests that, it is sometimes acceptable to add a detached variable into the loss. 2) Increasing the mask ratio does not necessarily lead to worse results, implying that OOD responses might not lead to performance degradation. 

\subsection{Experiment on LLMs}

We also experimented on LLM fine-tuning to further verify the effect of the method on more sophisticated cases. Due to the budget limitation, we train a Llama7b model on two datasets, SafeRLHF\cite{beavertails, safe-rlhf} and Anthropic HH\cite{bai2022hh}. For SafeRLHF, we use the official training script to get a reward model and a cost model, then evaluate the fine-tuned model with them. For Anthropic HH, as it has two subsets, namely 'helpful' and 'harmless', we fine-tune the model on both subsets, and we train a reward model for each subset individually to provide a more comprehensive evaluation.

\begin{table}[t]
    \centering
    \caption{Results on the SafeRLHF dataset. Higher \textbf{Help} and \textbf{S.R.}, and lower \textbf{Harm} are better.}
    \label{table:saferlhf}
    \resizebox{0.75\columnwidth}{!}{%
    \begin{tabular}{lccc}
        \toprule
        \textbf{Metric} & \textbf{Help} $\uparrow$ & \textbf{Harm} $\downarrow$ & \textbf{S.R.} $\uparrow$ \\
        \midrule
        DPO            & -0.69  & -1.94  & 68.2\%  \\
        nbDPO (asymm)  & -0.21  & -2.55  & 77.7\%  \\
        nbDPO (symm)   & -0.81  & -1.82  & 67.6\%  \\
        nbDPOv2 (asymm) & -0.22  & -2.49  & 77.5\%  \\
        nbDPOv2 (symm)  & -0.81  & -1.82  & 67.6\%  \\
        bDPO           & 0.78   & -4.13  & 86.2\%  \\
        \bottomrule
    \end{tabular}%
    }
\end{table}

Table~\ref{table:saferlhf} shows the result achieved on the SafeRLHF dataset. Our method shows a significant advantage compared with the baseline. It is worth noticing that, although symmetric nbDPO performed very well in the synthetic dataset, its performance is not satisfactory in real cases. We found in its training process that, the training process of symmetric nbDPO is very unstable in both cases, as the validation loss might sometimes increase a lot suddenly during training; in the synthetic dataset, this can be quickly recovered, but in real datasets, this becomes fatal.

\begin{table}[ht]
    \centering
    \caption{Result for Anthropic HH Dataset.}
    \label{table:hh}
    \resizebox{0.65\columnwidth}{!}{%
    \begin{tabular}{lccc}
        \toprule
        \textbf{Metric} & \textbf{Help} $\uparrow$ & \textbf{Harm} $\downarrow$ & \textbf{S.R.} $\uparrow$ \\
        \midrule
        baseline       & 1.38  & -2.98  & 78.3\%  \\
        bDPO           & 1.45  & -3.81  & 81.9\% \\
        \bottomrule
    \end{tabular}%
    }
\end{table}

Table~\ref{table:hh} shows the result achieved on the HH dataset. Although the two methods have comparable performance in helpfulness, our method shows a significant advantage in harmlessness. The reason why these two metrics have different results has not been explored yet.

\section{Discussion and Limitation} \label{section:discussion}

Because of limit of budgets, we did not conduct more thorough experiments on more real-world cases, on larger models or datasets (for example Ultrafeedback\cite{cui2023ultrafeedback} and CodeContest\cite{li2022codecontest}).

Moreover, as our analysis is based on discrete prompt and response space, future work might extend this theory to continuous cases and state the definition of 'out-of-distribution' in those cases more clearly.

We also acknowledge that, our methods might not be the SOTA methods on these datasets, but instead of achieving a new SOTA, what we want more is to prove with this method that diminishing the gradient imbalance is a promising way to finding a better algorithm for DPO, and that apart from the properties of the optimal solutions we should also focus on how the model is updated throughout the learning process.

\section{Conclusion} \label{section:conclusion}

Our work finds that, the unsatisfactory performance of DPO in some scenarios is caused by the imbalanced gradient in its loss function. We make a thorough analysis of the impact of gradient imbalance and also propose a novel method, Balanced-DPO, to mitigate the problem. We wish to demonstrate with our analysis and method that: 1) the gradient imbalance in DPO is a crucial issue hindering its performance, and 2) it is beneficial to take a glance at the model update during the training process besides the optimal solutions for the loss function. We hope that future works will propose better methods to diminish the imbalance of gradient in DPO loss to achieve comparable performance and robustness as PPO.

\section*{Impact Statement}

This paper presents work whose goal is to advance the field of Machine Learning (or more specifically, about improving post-training of LLMs). There are many potential societal consequences of our work, none of which we feel must be specifically highlighted here.

\bibliography{reference}

\begin{thebibliography}{28}
\providecommand{\natexlab}[1]{#1}
\providecommand{\url}[1]{\texttt{#1}}
\expandafter\ifx\csname urlstyle\endcsname\relax
  \providecommand{\doi}[1]{doi: #1}\else
  \providecommand{\doi}{doi: \begingroup \urlstyle{rm}\Url}\fi

\bibitem[Agrawal et~al.(2024)Agrawal, Suzgun, Mackey, and Kalai]{agrawal2024languagemodelsknowtheyre}
Agrawal, A., Suzgun, M., Mackey, L., and Kalai, A.~T.
\newblock Do language models know when they're hallucinating references?, 2024.
\newblock URL \url{https://arxiv.org/abs/2305.18248}.

\bibitem[Azar et~al.(2023)Azar, Rowland, Piot, Guo, Calandriello, Valko, and Munos]{azar2023ipo}
Azar, M.~G., Rowland, M., Piot, B., Guo, D., Calandriello, D., Valko, M., and Munos, R.
\newblock A general theoretical paradigm to understand learning from human preferences, 2023.
\newblock URL \url{https://arxiv.org/abs/2310.12036}.

\bibitem[Brown et~al.(2020)Brown, Mann, Ryder, Subbiah, Kaplan, Dhariwal, Neelakantan, Shyam, Sastry, Askell, Agarwal, Herbert-Voss, Krueger, Henighan, Child, Ramesh, Ziegler, Wu, Winter, Hesse, Chen, Sigler, Litwin, Gray, Chess, Clark, Berner, McCandlish, Radford, Sutskever, and Amodei]{llmfewshotlearner}
Brown, T., Mann, B., Ryder, N., Subbiah, M., Kaplan, J.~D., Dhariwal, P., Neelakantan, A., Shyam, P., Sastry, G., Askell, A., Agarwal, S., Herbert-Voss, A., Krueger, G., Henighan, T., Child, R., Ramesh, A., Ziegler, D., Wu, J., Winter, C., Hesse, C., Chen, M., Sigler, E., Litwin, M., Gray, S., Chess, B., Clark, J., Berner, C., McCandlish, S., Radford, A., Sutskever, I., and Amodei, D.
\newblock Language models are few-shot learners.
\newblock In Larochelle, H., Ranzato, M., Hadsell, R., Balcan, M., and Lin, H. (eds.), \emph{Advances in Neural Information Processing Systems}, volume~33, pp.\  1877--1901. Curran Associates, Inc., 2020.
\newblock URL \url{https://proceedings.neurips.cc/paper_files/paper/2020/file/1457c0d6bfcb4967418bfb8ac142f64a-Paper.pdf}.

\bibitem[Chowdhury et~al.(2024)Chowdhury, Kini, and Natarajan]{chowdhury2024rdpo}
Chowdhury, S.~R., Kini, A., and Natarajan, N.
\newblock Provably robust dpo: Aligning language models with noisy feedback, 2024.
\newblock URL \url{https://arxiv.org/abs/2403.00409}.

\bibitem[Cui et~al.(2023)Cui, Yuan, Ding, Yao, Zhu, Ni, Xie, Liu, and Sun]{cui2023ultrafeedback}
Cui, G., Yuan, L., Ding, N., Yao, G., Zhu, W., Ni, Y., Xie, G., Liu, Z., and Sun, M.
\newblock Ultrafeedback: Boosting language models with high-quality feedback.
\newblock 2023.

\bibitem[Dai et~al.(2024)Dai, Pan, Sun, Ji, Xu, Liu, Wang, and Yang]{safe-rlhf}
Dai, J., Pan, X., Sun, R., Ji, J., Xu, X., Liu, M., Wang, Y., and Yang, Y.
\newblock Safe rlhf: Safe reinforcement learning from human feedback.
\newblock In \emph{The Twelfth International Conference on Learning Representations}, 2024.
\newblock URL \url{https://openreview.net/forum?id=TyFrPOKYXw}.

\bibitem[et~al.(2023{\natexlab{a}})]{touvron2023llama2openfoundation}
et~al., H.~T.
\newblock Llama 2: Open foundation and fine-tuned chat models, 2023{\natexlab{a}}.
\newblock URL \url{https://arxiv.org/abs/2307.09288}.

\bibitem[et~al.(2023{\natexlab{b}})]{anil2023palm2technicalreport}
et~al., R.~A.
\newblock Palm 2 technical report, 2023{\natexlab{b}}.
\newblock URL \url{https://arxiv.org/abs/2305.10403}.

\bibitem[et~al.(2022)]{bai2022hh}
et~al., Y.~B.
\newblock Training a helpful and harmless assistant with reinforcement learning from human feedback, 2022.
\newblock URL \url{https://arxiv.org/abs/2204.05862}.

\bibitem[Ethayarajh et~al.(2024)Ethayarajh, Xu, Muennighoff, Jurafsky, and Kiela]{ethayarajh2024kto}
Ethayarajh, K., Xu, W., Muennighoff, N., Jurafsky, D., and Kiela, D.
\newblock Kto: Model alignment as prospect theoretic optimization, 2024.
\newblock URL \url{https://arxiv.org/abs/2402.01306}.

\bibitem[Ji et~al.(2023)Ji, Liu, Dai, Pan, Zhang, Bian, Chen, Sun, Wang, and Yang]{beavertails}
Ji, J., Liu, M., Dai, J., Pan, X., Zhang, C., Bian, C., Chen, B., Sun, R., Wang, Y., and Yang, Y.
\newblock Beavertails: Towards improved safety alignment of {LLM} via a human-preference dataset.
\newblock In \emph{Thirty-seventh Conference on Neural Information Processing Systems Datasets and Benchmarks Track}, 2023.
\newblock URL \url{https://openreview.net/forum?id=g0QovXbFw3}.

\bibitem[Kadavath et~al.(2022)Kadavath, Conerly, Askell, Henighan, Drain, Perez, Schiefer, Hatfield-Dodds, DasSarma, Tran-Johnson, Johnston, El-Showk, Jones, Elhage, Hume, Chen, Bai, Bowman, Fort, Ganguli, Hernandez, Jacobson, Kernion, Kravec, Lovitt, Ndousse, Olsson, Ringer, Amodei, Brown, Clark, Joseph, Mann, McCandlish, Olah, and Kaplan]{kadavath2022languagemodelsmostlyknow}
Kadavath, S., Conerly, T., Askell, A., Henighan, T., Drain, D., Perez, E., Schiefer, N., Hatfield-Dodds, Z., DasSarma, N., Tran-Johnson, E., Johnston, S., El-Showk, S., Jones, A., Elhage, N., Hume, T., Chen, A., Bai, Y., Bowman, S., Fort, S., Ganguli, D., Hernandez, D., Jacobson, J., Kernion, J., Kravec, S., Lovitt, L., Ndousse, K., Olsson, C., Ringer, S., Amodei, D., Brown, T., Clark, J., Joseph, N., Mann, B., McCandlish, S., Olah, C., and Kaplan, J.
\newblock Language models (mostly) know what they know, 2022.
\newblock URL \url{https://arxiv.org/abs/2207.05221}.

\bibitem[Konda \& Tsitsiklis(1999)Konda and Tsitsiklis]{konda1999actor}
Konda, V. and Tsitsiklis, J.
\newblock Actor-critic algorithms.
\newblock \emph{Advances in neural information processing systems}, 12, 1999.

\bibitem[Lee et~al.(2024)Lee, Bai, Pres, Wattenberg, Kummerfeld, and Mihalcea]{lee2024toxic}
Lee, A., Bai, X., Pres, I., Wattenberg, M., Kummerfeld, J.~K., and Mihalcea, R.
\newblock A mechanistic understanding of alignment algorithms: A case study on dpo and toxicity.
\newblock \emph{arXiv preprint arXiv:2401.01967}, 2024.

\bibitem[Li et~al.(2022)Li, Choi, Chung, Kushman, Schrittwieser, Leblond, Eccles, Keeling, Gimeno, Dal~Lago, et~al.]{li2022codecontest}
Li, Y., Choi, D., Chung, J., Kushman, N., Schrittwieser, J., Leblond, R., Eccles, T., Keeling, J., Gimeno, F., Dal~Lago, A., et~al.
\newblock Competition-level code generation with alphacode.
\newblock \emph{Science}, 378\penalty0 (6624):\penalty0 1092--1097, 2022.

\bibitem[OpenAI(2024)]{openai2024gpt4technicalreport}
OpenAI.
\newblock Gpt-4 technical report, 2024.
\newblock URL \url{https://arxiv.org/abs/2303.08774}.

\bibitem[Ouyang et~al.(2022)Ouyang, Wu, Jiang, Almeida, Wainwright, Mishkin, Zhang, Agarwal, Slama, Gray, Schulman, Hilton, Kelton, Miller, Simens, Askell, Welinder, Christiano, Leike, and Lowe]{ouyang2022traininglmintruction}
Ouyang, L., Wu, J., Jiang, X., Almeida, D., Wainwright, C., Mishkin, P., Zhang, C., Agarwal, S., Slama, K., Gray, A., Schulman, J., Hilton, J., Kelton, F., Miller, L., Simens, M., Askell, A., Welinder, P., Christiano, P., Leike, J., and Lowe, R.
\newblock Training language models to follow instructions with human feedback.
\newblock In Oh, A.~H., Agarwal, A., Belgrave, D., and Cho, K. (eds.), \emph{Advances in Neural Information Processing Systems}, 2022.
\newblock URL \url{https://openreview.net/forum?id=TG8KACxEON}.

\bibitem[Rafailov et~al.(2024)Rafailov, Sharma, Mitchell, Manning, Ermon, and Finn]{rafailov2024direct}
Rafailov, R., Sharma, A., Mitchell, E., Manning, C.~D., Ermon, S., and Finn, C.
\newblock Direct preference optimization: Your language model is secretly a reward model.
\newblock \emph{Advances in Neural Information Processing Systems}, 36, 2024.

\bibitem[Saeidi et~al.(2024)Saeidi, Verma, and Baral]{saeidi2024insights}
Saeidi, A., Verma, S., and Baral, C.
\newblock Insights into alignment: Evaluating dpo and its variants across multiple tasks.
\newblock \emph{arXiv preprint arXiv:2404.14723}, 2024.

\bibitem[Sanghi(2024)]{Sanghi2024}
Sanghi, N.
\newblock \emph{Proximal Policy Optimization (PPO) and RLHF}, pp.\  461--522.
\newblock Apress, Berkeley, CA, 2024.
\newblock ISBN 979-8-8688-0273-7.
\newblock \doi{10.1007/979-8-8688-0273-7_11}.
\newblock URL \url{https://doi.org/10.1007/979-8-8688-0273-7_11}.

\bibitem[Stiennon et~al.(2020)Stiennon, Ouyang, Wu, Ziegler, Lowe, Voss, Radford, Amodei, and Christiano]{stiennon2020summarizewithhumanfeedback}
Stiennon, N., Ouyang, L., Wu, J., Ziegler, D., Lowe, R., Voss, C., Radford, A., Amodei, D., and Christiano, P.~F.
\newblock Learning to summarize with human feedback.
\newblock In Larochelle, H., Ranzato, M., Hadsell, R., Balcan, M., and Lin, H. (eds.), \emph{Advances in Neural Information Processing Systems}, volume~33, pp.\  3008--3021. Curran Associates, Inc., 2020.
\newblock URL \url{https://proceedings.neurips.cc/paper_files/paper/2020/file/1f89885d556929e98d3ef9b86448f951-Paper.pdf}.

\bibitem[Wu et~al.(2024{\natexlab{a}})Wu, Wang, Yang, Wu, Gao, Ding, Wang, and He]{wu2024alpha}
Wu, J., Wang, X., Yang, Z., Wu, J., Gao, J., Ding, B., Wang, X., and He, X.
\newblock $\alpha$-dpo: Adaptive reward margin is what direct preference optimization needs.
\newblock \emph{arXiv preprint arXiv:2410.10148}, 2024{\natexlab{a}}.

\bibitem[Wu et~al.(2024{\natexlab{b}})Wu, Xie, Yang, Wu, Gao, Ding, Wang, and He]{wu2024beta}
Wu, J., Xie, Y., Yang, Z., Wu, J., Gao, J., Ding, B., Wang, X., and He, X.
\newblock $\beta$-dpo: Direct preference optimization with dynamic $\beta$.
\newblock \emph{arXiv preprint arXiv:2407.08639}, 2024{\natexlab{b}}.

\bibitem[Xiong et~al.(2024)Xiong, Dong, Ye, Wang, Zhong, Ji, Jiang, and Zhang]{xiong2024iterative}
Xiong, W., Dong, H., Ye, C., Wang, Z., Zhong, H., Ji, H., Jiang, N., and Zhang, T.
\newblock Iterative preference learning from human feedback: Bridging theory and practice for rlhf under kl-constraint.
\newblock In \emph{Forty-first International Conference on Machine Learning}, 2024.

\bibitem[Xu et~al.(2024{\natexlab{a}})Xu, Sharaf, Chen, Tan, Shen, Durme, Murray, and Kim]{xu2024cpo}
Xu, H., Sharaf, A., Chen, Y., Tan, W., Shen, L., Durme, B.~V., Murray, K., and Kim, Y.~J.
\newblock Contrastive preference optimization: Pushing the boundaries of llm performance in machine translation, 2024{\natexlab{a}}.
\newblock URL \url{https://arxiv.org/abs/2401.08417}.

\bibitem[Xu et~al.(2024{\natexlab{b}})Xu, Fu, Gao, Ye, Liu, Mei, Wang, Yu, and Wu]{xu2024dposuperiorppollm}
Xu, S., Fu, W., Gao, J., Ye, W., Liu, W., Mei, Z., Wang, G., Yu, C., and Wu, Y.
\newblock Is dpo superior to ppo for llm alignment? a comprehensive study, 2024{\natexlab{b}}.
\newblock URL \url{https://arxiv.org/abs/2404.10719}.

\bibitem[Zheng et~al.(2023)Zheng, Dou, Gao, Hua, Shen, Wang, Liu, Jin, Liu, Zhou, Xiong, Chen, Xi, Xu, Lai, Zhu, Chang, Yin, Weng, Cheng, Huang, Sun, Yan, Gui, Zhang, Qiu, and Huang]{zheng2023secretsrlhflargelanguage}
Zheng, R., Dou, S., Gao, S., Hua, Y., Shen, W., Wang, B., Liu, Y., Jin, S., Liu, Q., Zhou, Y., Xiong, L., Chen, L., Xi, Z., Xu, N., Lai, W., Zhu, M., Chang, C., Yin, Z., Weng, R., Cheng, W., Huang, H., Sun, T., Yan, H., Gui, T., Zhang, Q., Qiu, X., and Huang, X.
\newblock Secrets of rlhf in large language models part i: Ppo, 2023.
\newblock URL \url{https://arxiv.org/abs/2307.04964}.

\bibitem[Ziegler et~al.(2020)Ziegler, Stiennon, Wu, Brown, Radford, Amodei, Christiano, and Irving]{ziegler2020finetuninglanguagemodelshuman}
Ziegler, D.~M., Stiennon, N., Wu, J., Brown, T.~B., Radford, A., Amodei, D., Christiano, P., and Irving, G.
\newblock Fine-tuning language models from human preferences, 2020.
\newblock URL \url{https://arxiv.org/abs/1909.08593}.

\end{thebibliography}
\bibliographystyle{icml2024}

\newpage
\appendix
\onecolumn

\section{Toy Experiment Result}
\begin{figure}
    \centering
    \includegraphics[width=0.8\columnwidth]{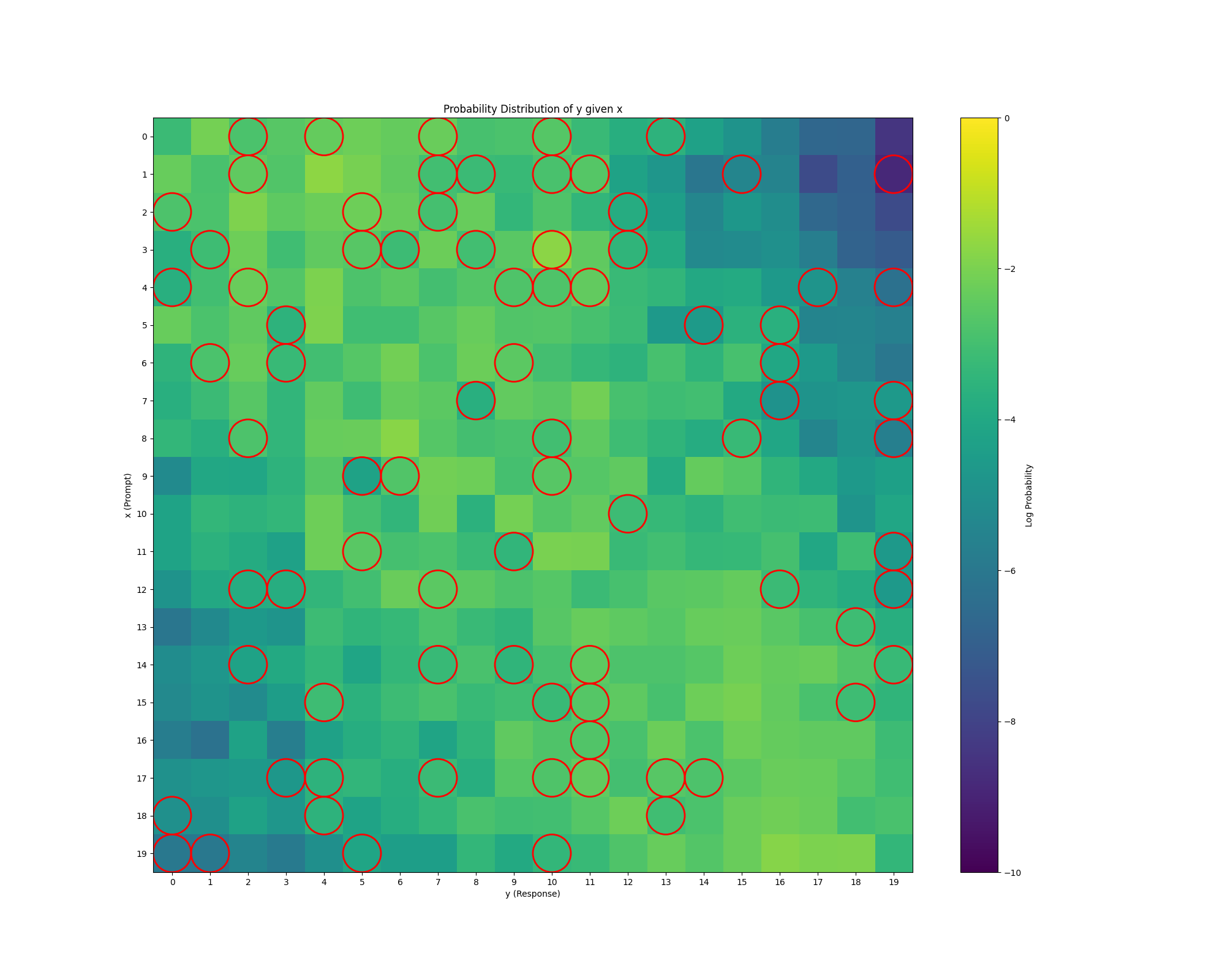}
    \caption{Probability distribution of the model output for DPO, where the lighter color indicates a higher probability and vice versa. This result is from the $mask=0.2$ case.}
    \label{fig:vis_dpooodresult}
\end{figure}

Figure~\ref{fig:vis_dpooodresult} shows the probability distribution of the model output for DPO, where the lighter color indicates a higher probability. 

\section{Experiment setting}

In the synthetic experiment, we let $\alpha=0.6$, $clip=0.5$ for nbDPO, and $\beta=1$ for all algorithms. In order to have a reasonable reference model, we first run 300 steps with the baseline DPO algorithm to achieve the reference model for other algorithms. For SafeRLHF, we set $clip=0.3$ for nbDPO, and the training process altogether has 2 epochs (the same for HH).

\section{Discussion for Negative Correlation} \label{appendix:negativecorr}

As we assume that $c_{ij} \approx 0$ in Assumption~\ref{assumption2}, we empirically find that $\mathbb{E}_{i,j}[c_{ij}]<0$ as stated in Section~\ref{section:assumptions}. We make a brief analysis on how this would impact our proof.

Let $\lambda=-\mathbb{E}[c_{ij}]$, then for DPO:
\begin{equation}
    \mathbb{E}[\Delta^js_i] = (1-\lambda) \eta g^2\tau_{ij}(1-\kappa_ij)
\end{equation}
The $\lambda$ term will decrease $s_i$ if $y_i$ is the winning response, and vice versa. This is a not expected behavior.
On the contrary, for a balanced loss:
\begin{equation}
    \Delta^js_i = C\eta g^2\tau_{ij}(1-\kappa_{ij})(1-\probi+\probj + \lambda(\probi-\probi^2+\probi\probj-2\probj]    
\end{equation}
The $\lambda$ term will increase $s_i$ for both responses. Generally, the $s_i$ for the losing response will be increased more than the winning response, Hence, the negative correlation will make DPO have less advantage against a balanced loss, making the issue more severe.

\section{Omitted Proofs}

\paragraph{Lemma~\ref{lemma:pracps} Proof}
\begin{proof}
    It is an obvious result from Equation~\ref{eq:logitandprob}.
\end{proof}

\paragraph{Theorem~\ref{theorem:probupdate} Proof}

\begin{proof}
    Let's first prove the case for DPO. We follow the symbols in Section~\ref{section:oodresponses} here.
    \begin{align}
        \Delta s_i &= \sum\limits_{(i, j, \tau_{ij})\in \mathcal{A}}\Delta^js_i \\
        &= \gamma w_i
    \end{align}
    
    And the probability update $\Delta p_i$ of a response $(i\in\mathcal{A})$ should be:
    \begin{align}
        \Delta p_i &= \sum\limits_{j\in A}\prac{p_i}{s_j}\Delta s_j \\
        &= \gamma w_ip_i(1-p_i) - \sum\limits_{j\neq i}\gamma w_jp_ip_j \\
        &= \gamma p_i (w_i - \sum\limits_{j\in A}w_jp_j)
    \end{align}

    As for a balanced loss, we can see that $\Delta_b^js_i = C\probi\Delta_d^js_i$, where $\Delta_b^js_i$ denotes $\Delta^js_i$ for a balanced loss, and $\Delta_d^js_i$ denotes this for DPO.
    Hence, this is easily extended from the form in DPO.
\end{proof}

\paragraph{Theorem~\ref{theorem:distshift} Proof}
\begin{proof}  
    By Lemma~\ref{lemma:wi_simplify}, 
    Let $q_i=e^{r^*(x,y_i)}$, If $r^*(x,y)$ decays super-linearly, then $q_i$ decays super-exponentially. 
    \begin{equation}
        w(y_i) = -\sum_{j\neq i}(\frac{p_i}{p_i+p_j}-\frac{q_i}{q_i+q_j})
    \end{equation}
    \begin{equation}
        \prac{w(y_i)}{y_i} =  - (\frac{p_j p_i'}{(p_i+p_j)^2} - \frac{q_j q_i'}{(q_i+q_j)^2})
    \end{equation}
    Since both $p_i$ and $q_i$ decays super-exponentially, we can see that
    \begin{align}
        p_i' &= -f_p(y-\mu_p) p_i \\
        q_i' &= -f_q(y-\mu_q) q_i
    \end{align}
    where $f_p(\cdot)$ and $f_q(\cdot)$ are strictly monotonically increasing functions satisfying $f_p(0)=f_q(0)=0$.

    Hence, $y^*$ should satisfy
    \begin{equation}
        \sum_{j\neq i}(f_p(y_i-\mu_p)\frac{p_ip_j}{(p_i+p_j)^2} - f_q(y_i-\mu_q)\frac{q_iq_j}{(q_i+q_j)^2}) = 0
    \end{equation}
    For any $\mu_p< y_i<\mu_q$, both terms are positive, hence the equation would not hold. If $y_i=\mu_p$ then the second term is positive while the first term is zero, and similar for $y_i=\mu_q$. Hence, the first statement is proved.

    For the second statement, let's consider the derivative of $\hat{w}(y_i)$. Similar to the analysis on $w(y_i)$, we can see that:

    \begin{equation}
        \prac{\hat{w}(y)}{y} = -\sum_{j\neq i}(f_p(y_i-\mu_p)\frac{p_i^2p_j}{(p_i+p_j)^2} + f_p(y_i-\mu_p)\frac{p_i^2}{p_i+p_j}- f_q(y_i-\mu_q)\frac{q_iq_j}{(q_i+q_j)^2}p_i - f_p(y_i-\mu_p)\frac{q_ip_i}{q_i+q_j})
    \end{equation}
    Obviously $p_i=0$ is invalid. Hence,
    \begin{equation}
        \label{eq:definitiong}
        \sum_{j\neq i}(f_p(y_i-\mu_p)\frac{p_ip_j}{(p_i+p_j)^2} + f_p(y_i-\mu_p)\frac{p_i}{p_i+p_j}- f_q(y_i-\mu_q)\frac{q_iq_j}{(q_i+q_j)^2} - f_p(y_i-\mu_p)\frac{q_i}{q_i+q_j}) = 0
    \end{equation}
    Let the left hand side be $g(y_i)$, then:
    \begin{equation}
        g(y_i) = -\prac{w(y)}{y} + f_p(y_i-\mu_p)\sum_{j\neq i}(\frac{p_i}{p_i+p_j}-\frac{q_i}{q_i+q_j}) 
    \end{equation}
    We can assume $y^*$ is the closest local extremum to $\mu_q$ that satisfies $y^*>\mu_q$ (obviously if this holds then for any $y^*>\mu_q$ the statement holds). Since $y^*$ satisfies
    \begin{equation}
        \left. \prac{w(y)}{y}\right|_{y=y^*} = 0
    \end{equation}
    Then
    \begin{equation}
        g(y^*) = -f_p(y^*-\mu_p)w(y^*)
    \end{equation}
    Since $w(\mu_q)>0$ and $\left. \prac{w(y)}{y} \right|_{y=\mu_q} > 0$, $w(y^*)>0$. Hence, $g(y^*)<0$. 
    On the other hand, it is easy to verify from the definition of $g(\cdot)$ (the left hand side of Equation~\ref{eq:definitiong}) that $g(\mu_q)>0$, and since $g$ is continuous, the second statement is proved.

    The third statement's proof is similar to the second. We also assume that $y^*$ is the closest local extremum of $w(y)$ to $\mu_p$. Since $w(\mu_p)<0$ and $\left. \prac{w(y)}{y} \right|_{y=\mu_q} > 0$, we can see $w(y^*)<0$, thus $g(y^*)<0$. However, as $g(\mu_p)>0$, the third statement is proved.    
\end{proof}

\paragraph{Lemma~\ref{lemma:wi_simplify} Proof}
\begin{proof}
    It is an obvious transformation of $w_i$ so we omit the proof here.
\end{proof}

\paragraph{Lemma~\ref{lemma:fym} Proof}
\begin{proof}
    \begin{align}
    & \operatorname{Var}[f(X) g(X)] \\
    = & \frac{1}{2} \mathbb{E}\left[(f(X) g(X)-f(Y) g(Y))^2\right] \\
    \geq & \frac{1}{2} \mathbb{E}\left[(f(X)-f(Y))^2 \max \{g(X), g(Y)\}^2\right] \\
    = & \mathbb{E}\left[\int_{-\infty}^{\infty} \int_{-\infty}^{\infty} \max \{g(X), g(Y)\}^2 \mathbb{1}\{f(X) \leq r, s \leq f(Y)\} \mathrm{d} r \mathrm{~d} s\right] \\
    = & \int_{-\infty}^{\infty} \int_{-\infty}^{\infty} \mathbb{E}\left[\max \{g(X), g(Y)\}^2 \mathbb{1}\{f(X) \leq r, s \leq f(Y)\}\right] \mathrm{d} r \mathrm{~d} s \\
    = & \int_{-\infty}^{\infty} \int_{-\infty}^{\infty} \mathbb{P}[f(X) \leq \min \{r, s\}] \cdot \mathbb{P}[f(X) \geq \max \{r, s\}] \cdot \mathbb{E}\left[g(X)^2 \mid f(X) \geq \max \{r, s\}\right] \mathrm{d} r \mathrm{~d} s \\
    \geq & \int_{-\infty}^{\infty} \int_{-\infty}^{\infty} \mathbb{P}[f(X) \leq \min \{r, s\}] \cdot \mathbb{P}[f(X) \geq \max \{r, s\}] \cdot \mathbb{E}\left[g(X)^2\right] \mathrm{d} r \mathrm{~d} s \\
    = & \mathbb{E}\left[g(X)^2\right] \int_{-\infty}^{\infty} \int_{-\infty}^{\infty} \mathbb{P}[f(X) \leq \min \{r, s\}] \cdot \mathbb{P}[f(X) \geq \max \{r, s\}] \cdot \mathrm{d} r \mathrm{~d} s \\
    = & \mathbb{E}\left[g(X)^2\right] \int_{-\infty}^{\infty} \int_{-\infty}^{\infty} \mathbb{E}[\mathbb{1}\{f(X) \leq r, s \leq f(Y)\}] \cdot \mathrm{d} r \mathrm{~d} s \\
    = & \mathbb{E}\left[g(X)^2\right] \cdot \mathbb{E}\left[\int_{-\infty}^{\infty} \int_{-\infty}^{\infty} \mathbb{1}\{f(X) \leq r, s \leq f(Y)\} \mathrm{d} r \mathrm{~d} s\right] \\
    = & \mathbb{E}\left[g(X)^2\right] \cdot \frac{1}{2} \mathbb{E}\left[(f(X)-f(Y))^2\right] \\
    = & \mathbb{E}\left[g(X)^2\right] \cdot \operatorname{Var}[f] 
    \end{align}
\end{proof}

\paragraph{Theorem~\ref{theorem:variance} Proof}

\begin{proof}
    \begin{align}
        \operatorname{Var}\left[w_ip_i\right] &= \operatorname{Var}[\alpha_ip_i] + \operatorname{Var}[\beta_ip_i] - 2\operatorname{Cov}[\alpha_ip_i, \beta_ip_i] \\
        &\geq (\operatorname{Var}[\alpha_i] + \operatorname{Var}[\beta_i])\mathbb{E}[p_i^2] - 2\operatorname{Cov}[\alpha_ip_i, \beta_ip_i]
    \end{align}
    \begin{align}
        \operatorname{Var}\left[w_ip_i\right] - \operatorname{Var}[w_i]\mathbb{E}[p_i^2] &\geq 2\operatorname{Cov}[\alpha_i,\beta_i]\mathbb{E}[p_i^2] - 2\operatorname{Cov}[\alpha_ip_i, \beta_ip_i]
    \end{align}
    Expanding the right hand side would achieve the result.
\end{proof}

\paragraph{Theorem~\ref{theorem:dataquality} Proof}

\begin{proof}
    \begin{align}
        (\operatorname{Var}[w_i]-\operatorname{Var}[w^*_i])\mathbb{E}_{y_i\sim \mathcal{A}
        }[p_i^2] &= (\mathbb{E}[(w_i^*+\epsilon_i)^2] - \mathbb{E}[(w_i^*)^2]) \mathbb{E}_{y_i\sim \mathcal{A}
        }[p_i^2] \\ 
        &\leq (\mathbb{E}[(w_i^*+\epsilon_i)^2] - \mathbb{E}[(w_i^*)^2])\mathbb{E}_{y_i\sim \mathcal{A}_0
        }[p_i^2] \\
        &= \mathbb{E}[\epsilon_i^2]\mathbb{E}[p_i^2]
    \end{align}
    \begin{align}
         \operatorname{Var}[w_ip_i] - \operatorname{Var}[w^*_ip_i] 
         &= \mathbb{E}[(w_i^*+\epsilon_i)^2p_i^2] - \mathbb{E}[(w_i^*p_i)^2] - (\mathbb{E}[(w_i^*+\epsilon_i)p_i])^2 + (\mathbb{E}[w_i^*p_i])^2 \\
         &= \mathbb{E}[\epsilon_i^2]\mathbb{E}[p_i^2]
    \end{align}
    Combining the above formulas, statement proved.
\end{proof}

\paragraph{Theorem~\ref{theorem:oodnoprob} Proof}

\begin{proof}
    By Theorem~\ref{theorem:probupdate}, if $\probi=0$ then $\Delta \probi=0$. Hence it is obvious.
\end{proof}

\paragraph{Lemma~\ref{lemma:fym2} Proof}

\begin{proof}
    \begin{align}
    & \operatorname{Cov}[f(X) g(X), g(X)] \\
    = & \frac{1}{2} \mathbb{E}[(f(X) g(X)-f(Y) g(Y))(g(X)-g(Y))] \\
    \geq & \frac{1}{2} \mathbb{E}\left[(g(X)-g(Y))^2 \max \{f(X), f(Y)\}\right] \\
    = & \mathbb{E}\left[\int_{-\infty}^{\infty} \int_{-\infty}^{\infty} \max \{f(X), f(Y)\} \mathbb{1}\{g(X) \leq r, s \leq g(Y)\} \mathrm{d} r \mathrm{~d} s\right] \\
    = & \int_{-\infty}^{\infty} \int_{-\infty}^{\infty} \mathbb{E}[\max \{f(X), f(Y)\} \mathbb{1}\{g(X) \leq r, s \leq g(Y)\}] \mathrm{d} r \mathrm{~d} s \\
    = & \int_{-\infty}^{\infty} \int_{-\infty}^{\infty} \mathbb{P}[g(X) \leq \min \{r, s\}] \cdot \mathbb{P}[g(X) \geq \max \{r, s\}] \cdot \mathbb{E}[f(X) \mid g(X) \geq \max \{r, s\}] \mathrm{d} r \mathrm{~d} s \\
    \geq & \int_{-\infty}^{\infty} \int_{-\infty}^{\infty} \mathbb{P}[g(X) \leq \min \{r, s\}] \cdot \mathbb{P}[g(X) \geq \max \{r, s\}] \cdot \mathbb{E}[f(X)] \mathrm{d} r \mathrm{~d} s \\
    = & \mathbb{E}[f(X)] \int_{-\infty}^{\infty} \int_{-\infty}^{\infty} \mathbb{P}[g(X) \leq \min \{r, s\}] \cdot \mathbb{P}[g(X) \geq \max \{r, s\}] \cdot \mathrm{d} r \mathrm{~d} s \\
    = & \mathbb{E}[f(X)] \int_{-\infty}^{\infty} \int_{-\infty}^{\infty} \mathbb{E}[\mathbb{1}\{g(X) \leq r, s \leq g(Y)\}] \cdot \mathrm{d} r \mathrm{~d} s \\
    = & \mathbb{E}[f(X)] \cdot \mathbb{E}\left[\int_{-\infty}^{\infty} \int_{-\infty}^{\infty} \mathbb{1}\{g(X) \leq r, s \leq g(Y)\} \mathrm{d} r \mathrm{~d} s\right] \\
    = & \mathbb{E}[f(X)] \cdot \frac{1}{2} \mathbb{E}\left[(g(X)-g(Y))^2\right] \\
    = & \mathbb{E}[f(X)] \cdot \operatorname{Var}[g(X)]
    \end{align}
\end{proof}

\paragraph{Theorem~\ref{theorem:oodupdate} Proof}

\begin{proof}
    It is equivalent to:
    \begin{equation}
        \mathbb{E}\left[w_ip_i^2\right] < \mathbb{E}\left[w_ip_i\right]\mathbb{E}\left[p_i\right]
    \end{equation}
    By Lemma~\ref{lemma:fym2} we can see that:
    \begin{align}
        \mathbb{E}\left[\alpha_ip_i(p_i-\mathbb{E}_{k\in\mathbb{A}}[p_k])\right] & \geq \mathbb{E}[\alpha_i](\mathbb{E}[p_i^2]-\mathbb{E}[p_i]^2) \\
         &= \mathbb{E}[\beta_i](\mathbb{E}[p_i^2]-\mathbb{E}[p_i]^2)
    \end{align}
    By Lemma~\ref{lemma:wi_simplify}, statement proved.
\end{proof}


\end{document}